\documentclass[]{spie}  %>>> use for US letter paper
 \usepackage{float}
\usepackage{amsmath,amsfonts,amssymb}
\usepackage{graphicx}
\usepackage[colorlinks=true, allcolors=blue]{hyperref}

\usepackage{multirow}
\usepackage{booktabs}

\title{Hybrid bundle-adjusting 3D Gaussians for view consistent rendering with pose optimization}

\author[a]{Yanan Guo}
\author[a]{Ying Xie}
\author[b]{Ying Chang}
\author[b,*]{Benkui Zhang}
\author[a]{Bo Jia}
\author[a]{Lin Cao}
\affil[a]{Key Laboratory of Information and Communication Systems, Ministry of Information Industry, Beijing Information Science and Technology University, Beijing, China}
\affil[b]{Key Laboratory of Target Cognition and Application Technology, Aerospace Information Research Institute, Beijing, China}

\authorinfo{*Address all correspondence to Benkui Zhang zhangbk0566@126.com}

\begin{document} 
\maketitle

\begin{abstract}
Novel view synthesis has made significant progress in the field of 3D computer vision. However, the rendering of view-consistent novel views from imperfect camera poses remains challenging. In this paper, we introduce a hybrid bundle-adjusting 3D Gaussians model that enables view-consistent rendering with pose optimization. This model jointly extract image-based and neural 3D representations to simultaneously generate view-consistent images and camera poses within forward-facing scenes. The effective of our model is demonstrated through extensive experiments conducted on both real and synthetic datasets. These experiments clearly illustrate that our model can effectively optimize neural scene representations while simultaneously resolving significant camera pose misalignments. The source code is available at https://github.com/Bistu3DV/hybridBA.
\end{abstract}

\keywords{novel view synthesis, view consistent rendering, hybrid bundle-adjusting 3D Gaussians, camera poses register}

\section{INTRODUCTION}
\label{sec:intro} 
Novel view synthesis refers to the task of rendering a target image corresponding to a target pose, given a source image and source pose, as well as the target pose.
Novel view synthesis is a long-standing challenge in the domain of 3D computer vision with applications in deep learning data augmentation, automatic driving, and robot navigation.

Recent methods have demonstrated powerful rendering effects on novel view synthesis. NeRF\cite{NeRF} and NeRF--\cite{NeRF--} optimise a continuous 5D neural radiation field representation of the scene from a set of input images using volume rendering in order to render the scene from any angle. 3D Gaussian Splatting (3DGS)\cite{3DGS} and Scaffold-GS\cite{scaffold-gs} are initialised with point cloud derived from the Structure from Motion (SfM)\cite{SfM}, optimising a set of 3D Gaussian distributions to represent the scene. However, these methods exhibit a high sensitivity to camera pose noise during training, and even small pose deviations can have a significant impact on the final rendered view.

To improve the quality of novel view synthesis, current research trends can be broadly categorized into two main methods. Techniques such as BARF\cite{BARF} and SPARF\cite{SPARF} have shown significant success in overcoming challenges related to imperfect camera pose inputs. However, these methods are computationally intensive, require substantial memory, and suffer from slow rendering speeds during the optimization of individual scenes. The other method, exemplified by Gaussian-barf\cite{FlowMap}, concentrates on optimizing 3DGS. This method offers a differentiable and straightforward method for estimating camera and geometric parameters. It utilizes pre-existing optical flow and point track correspondences to monitor and direct the optimization process, which is then refined with a scene-dependent gradient descent algorithm. The end result is the synthesis of highly realistic and novel views. However, this method suffers from the challenge of baking perspective-related effects into a single Gaussian parameter that lacks interpolation capabilities.
Consequently, the effectiveness of these methods is compromised under significant viewpoint variations and changes in lighting conditions. Our method introduces image features into Gaussian framework to achieve view consistent rendering under pose inaccuracy conditions, while jointly optimising the pose and model parameters to correct the camera pose.

In this paper, we introduce a hybrid bundle-adjusting 3D Gaussians model that enables view-consistent rendering with pose optimization to solve these challenges.
Firstly, to render view consistent novel images, our model initiates the rendering process by transforming the scene's point cloud into a voxelized format, thereby extracting neural 3D anchor features. As each anchor feature is projected onto a specific view, we further enhance the rendering process by extracting image features from nearby views. The features surrounding the projected anchor point location are instrumental in constructing the image-based features for view consistent rendering.
Subsequently, we employ a coarse-to-fine bundle adjustment technique to parameterize camera pose, which allows us to jointly reconstruct the 3D scene and register the camera poses. To ensure alignment of camera poses for test views prior to rendering, we preserve our model’s trained state on the training views, keeping it frozen while we register the camera poses for the test views. This optimization is achieved by minimizing the discrepancies between the synthesized images and the actual test views.
Finally, based on the registered camera pose, we refine the alignment between the image-based representation and the neural 3D representation to improve the overall quality of the rendered images.
The effective of our model is demonstrated through extensive experiments conducted on both real and synthetic datasets. These experiments clearly illustrate that our model can effectively optimize neural scene representations while simultaneously resolving significant camera pose misalignments.

While our model is built upon the state-of-the-art 3D and image-based neural rendering models, our contribution is mainly in imperfect camera pose calibration, which produces images and camera poses with a consistent view in the scene. 
Our main contributions are as follows:
1) In the case of imperfect camera pose, we introduce the features of nearby images and propose a method of fusing point cloud with image features;  
2) We employ a coarse-to-fine bundle adjustment technique to parameterize camera pose, keep the trained model and frozen the camera pose in the test view.

\section{RELATED WORK}
\label{sec:related work}

\subsection{Image-Based Rendering}
The task of novel view synthesis is to render images from perspectives that differ from the original input view. This process is often achieved through image-based rendering\cite{LightFR,free} techniques. Traditional methods\cite{DeepBF,ScalableII,FreeVS} generally involve selecting a number of similar images, altering them to align with the target viewpoint, and then combining them to produce the final image. PixelNeRF\cite{pixelNeRF} can be trained across various scenes to understand scene-specific priors, enabling it to synthesize novel views from a limited set of input views in a feed-forward fashion.
LIRF\cite{LIRF} gathers information from conical view cones to form rays and renders high-quality novel views at continuous-value scales using volume rendering.
However, these methods have challenges in view-consistent novel view synthesis, especially if the input camera position is not accurate.

\subsection{Joint Camera Pose Optimization}
Recently, the NeRF--\cite{NeRF--} has introduced an enhanced NeRF\cite{NeRF} that facilitates the learning and refinement of camera parameters. This innovation streamlines the training of NeRF for forward scenes by obviating the need for known or precomputed camera parameters, such as intrinsic settings and 6DoF poses.
SC-NeRF\cite{SC-NeRF} leverages information from input views to derive density and color data within a continuous space, thereby illustrating the potential for enhancing both camera intrinsic and extrinsic.
GARF\cite{GARF} employs an encoder-decoder architecture combined with a point-level, learnable multi-view feature fusion module to extract common attributes for novel view synthesis, effectively mitigating issues related to occlusion.
Nope-NeRF\cite{NoPe-NeRF} incorporates an external monocular depth prediction model to refine camera positioning. The underlying concept of PoRF\cite{PoRF} is to utilize a single MLP to optimize the camera pose across all images within the dataset.
However, each of these methods suffers from the shortcomings of optimising only the positions or model parameters.

\section{METHOD}
\label{sec:method}

\subsection{Overview}
\label{sub:Overview}
\begin{figure} [ht]
   \begin{center}
   \begin{tabular}{c} 
   \includegraphics[width=\textwidth,keepaspectratio]{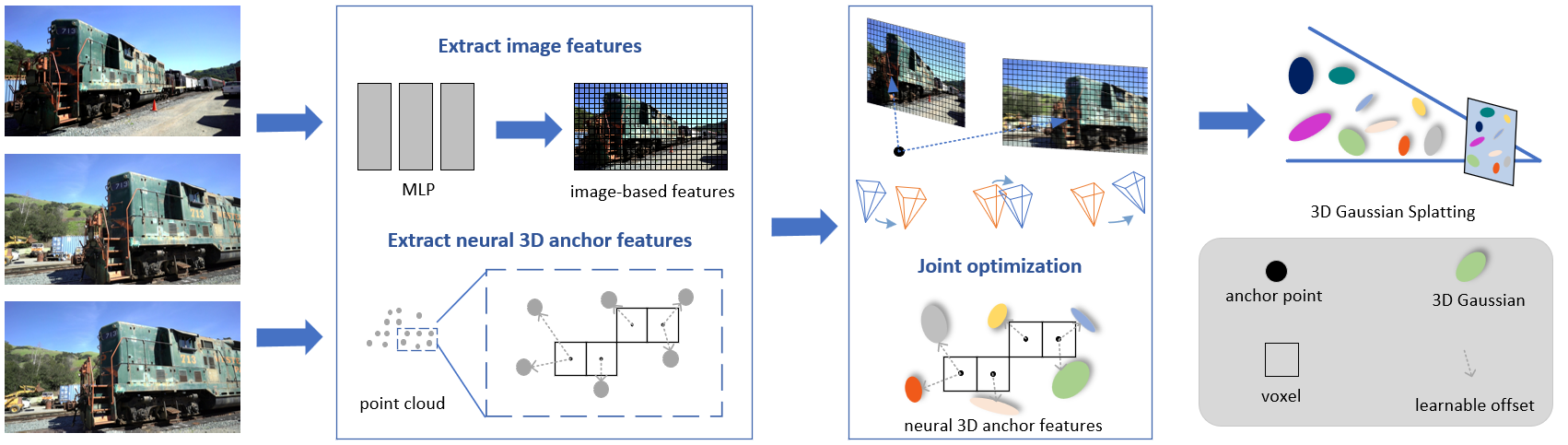}
  
   \end{tabular}
   \end{center}
   \caption[example] 
%>>>> use \label inside caption to get Fig. number with \ref{}
   { \label{fig:pipeline} 
\textbf{Overview of the method.} We first input a set of nearby views to extract pixel-level features for each image as well as key anchor point features in the scene. Next, we accurately match these image features with the 3D anchor points generated through the point cloud, which in turn enables the fusion of the image features with the anchor point features. In order to improve the accuracy, we correct the camera's position through joint optimisation, which ensures the accuracy of the camera parameters. Finally, we render the corrected data using 3DGS to synthesise novel views with high quality.}
   \end{figure} 

Given $n$ nearby visible light images ${\{{I_{1}},{I_{2}},...,{I_{n}}\}}$ as input to our model, our task is to perform a novel view synthesis of the scene and jointly optimise the camera poses and parameters. As shown in Fig.~\ref{fig:pipeline}, our model contains three main parts which are neural feature extraction, neural feature matching and joint optimisation of the position.
The neural feature extraction module is divided into two parts. On the one hand, we use MLP to extract image-based features $g_i^j$ pixel by pixel for the input images. On the other hand, we use SfM\cite{SfM} on the input images to generate sparse point cloud and divide the voxel grid ${V}$ to extract anchor points ${v}$.
On the neural feature matching module, we perform pixel-by-pixel matching of 3D anchor features with 2D image features to obtain 3D anchor points with hybrid image features $h_i$.
On the joint optimisation of the position module, we introduce a self-correction mechanism to jointly optimise for noisy anchor points and inaccurate camera bit positions, and render novel views via 3DGS\cite{3DGS}.

\subsection{Neural Feature Extraction}
Our model renders novel views by converting scene point cloud to voxels and extracting 3D anchor point features to ensure view consistency. At the same time, it uses the image features of nearby views to enhance rendering and achieve view consistency.
As shown in the Fig.~\ref{fig:pipeline}, we assign neural 3D features and image-based features for each 3D Gaussian.

\subsubsection{Image-based Features}
We employ a lightweight MLP to extract multi-scale image features ${\{{I_{1}^{j}},{I_{2}^{j}},...,{I_{n}^{j}}\}}$. These features are critical for subsequent processing steps because they provide detailed information about the image, allowing us to make the most of that information during the rendering process.

As shown in Eq.\ref{eq:c_i^j}, we take into account not only color ${c_i^j}$, but also the deviations of view directions ${\Delta d_i^j}$. The deviations of view directions refers to the difference in the perception of distance between the observer and the viewer when viewing the image due to the difference in viewing angle. By adding the deviations of view directions ${\Delta d_i^j}$ to image-based features ${\{{I_{1}^{j}},{I_{2}^{j}},...,{I_{n}^{j}}\}}$, we are able to make the rendered results more human-friendly, increasing the realism of the rendered effect.
\begin{equation}
  \label{eq:c_i^j}
  {c_i^j}=F_c({I_{i}^{j}}),{\Delta d_i^j}=F_d({I_{i}^{j}})
\end{equation}
where $F$ is a function that extracts features from the images.
Therefore, the image-based features are denoted by ${g_i^j=\{g_i^1,g_i^2,...,g_i^m\}}$, which is formed by integrating color ${c_i^j}$ and the deviations of view directions ${\Delta d_i^j}$ as follows:

\begin{equation}
  \label{eq:g_i^j}
  g_i^j=G({c_i^j},{\Delta d_i^j})
\end{equation}
where the function $G$ incorporates both ${c_i^j}$ and ${\Delta d_i^j}$ into the image-based features $g_i^j$.

\subsubsection{Anchor Point Features} 
Our method uses the point cloud ${P}\in\mathbb{R}$ generated by SfM\cite{SfM} as input. On this basis, we further divide this point cloud scene into sparse voxel mesh ${V}$ for more efficient processing and analysis of the point cloud. ${V}$ is defined as follows:
\begin{equation}
  \label{eq:V}
  {V}=\left\{\left\lfloor\frac{{P}}{\varepsilon }\right\rceil\right\}\cdot\varepsilon 
\end{equation}
where $\mathbf{\varepsilon}$ is the voxel scale.

After dividing the 3D space into a series of square voxel meshes ${V}$, we designate the position of the centre of each voxel as the anchor point of that voxel and denote it by the variable ${v}\in{V}$. The anchor points will be used for the extraction and generalisation of the surrounding point cloud. Specifically, these anchor points ${v}$ will be used as basic units for subsequent operations to extract and generalise feature information from the surrounding point cloud.

\subsection{Neural Feature Matching}
In this section, we will expound upon the processing for precisely aligning anchor points with image features. Specifically, our method involves projecting the 2D image features into the 3D world coordinate system. By calculating the minimal distance between each anchor point and the respective pixels, we ascertain the corresponding pixel position for each anchor point, thereby facilitating an efficient matching of neural features.

Since images are defined in 2D space and these images have corresponding camera extrinsic, we can use the inverse matrix of these camera extrinsic $\begin{bmatrix}
  R_c&C
\end{bmatrix}_{3\times 4} $ to transform the camera coordinates $P_{c}^c$, which are originally in 2D space, to 3D space. In this way, we can obtain the specific position of the camera corresponding to each image in the world coordinate system in 3D space, denoted as $P_{c}^w$ as follows:

\begin{equation}
  \label{eq:P_w}
  P_{c}^w=\begin{bmatrix}
  R_c&C
\end{bmatrix} \begin{bmatrix}
 P_{c}^c\\1

\end{bmatrix}
\end{equation}
where $\begin{bmatrix}
  R_c&C
\end{bmatrix}_{3\times 4} $ is the inverse matrix of the camera extrinsic, also known as the camera-to-world (c2w) matrix, used to transform coordinates from the camera coordinate system to the world coordinate system. $R_c$ is the rotation matrix and $C$ is the translation vector.

Given the coordinates $P_{a}^w$ of the anchor point in the world coordinate system and the camera position $P_{c}^w$, the relative distance $\delta_{ac}$ between them is calculated as shown in Eq.\ref{eq:delta}. We are able to calculate the minimum distance $\delta_{ac}'$ from each anchor point to the image, from which it is possible to determine which specific image corresponds to each anchor point.
\begin{equation}
  \label{eq:delta}
  \delta_{ac}=\|{P_{a}^w}-{P_{c}^w}\|_2,\delta_{ac}'=min(\delta_{ac})
  \end{equation}

Similar to the method described above, we map the image pixel by pixel into 3D space using the c2w matrix such that each pixel point in the image corresponds to a specific coordinate in the world coordinate system, denoted as $P_{p}^w$ as shown in Eq.\ref{eq:deltaap}. Next, we calculate the distance between each anchor point and its corresponding pixel points in the image and select the minimum distance $\delta_{ap}'$ from it, and the pixel point corresponding to $\delta_{ap}'$ is the matching pixel of the anchor point.

\begin{equation}
  \label{eq:deltaap}
  \delta_{ap}=\|{P_{a}^w}-{P_{p}^w}\|_2,\delta_{ap}'=min(\delta_{ap})
  \end{equation}

By means of this process, we achieve accurate match of neural features to obtain hybrid neural features $h_i$. The matching process involves the precise pairing of anchor points with pixel, thereby not only ensuring the accuracy of the positional calibration but also enhancing the information content of the point cloud by incorporating image features into the anchor points to compensate for the absence of inherent point cloud features.

\subsection{Optimization and Loss}
\subsubsection{Neural Gaussian Derivation}
We parameterise the neural Gaussian as position $\mathbf{\mu\in\mathbb{R}}$, opacity $\mathbf{\alpha\in\mathbb{R}}$, covariance-dependent quaternion ${q\in\mathbb{R}}$, scale ${s\in\mathbb{R}}$, and color ${c\in\mathbb{R}}$. A series of linear transformations and activation functions are used to decode these neural Gaussian properties from the hybrid neural features. Each of these transformations corresponds to a property of the neural Gaussian. Specifically, the opacity $\mathbf{\alpha}$ generated from the hybrid neural features can be computed using a specific formula:
\begin{equation}
  \label{eq:alpha}
\{\alpha_0,...,\alpha_{k-1}\}=E_\alpha(h_i,\delta_{ap})
\end{equation}

Similarly, we can derive the covariance-related quaternion ${E_q}$, scale ${E_s}$, and color ${E_c}$ of the neural Gaussian in a similar way. The core idea of this process is to break down complex properties into simple elements, and then learn the relationships between them through neural networks. It is worth noting that the prediction of neural Gaussian attributes in our method is dynamic. This way of dynamic activation greatly improves the efficiency of the algorithm because it avoids unnecessary calculations.

\subsubsection{Joint Optimization for Alignment}
In order to improve the accuracy and efficiency of registration, we adopted a coarse-to-fine bundle-adjusting registration strategy to optimize the pose. We first initialise a Gaussian distribution model based on the collected anchor data, and in the process employ the heuristic rules described in detail in 3DGS. In order to solve a series of problems due to inaccuracies in the camera position, we specifically introduce a self-correction mechanism. This mechanism efficiently adjusts the camera parameters and the parameters of the Gaussian distribution by utilising the photometric loss function, thus effectively improving the robustness and accuracy of the whole system. In this way, we are able to better adapt to the uncertainties in the camera imaging process and optimise the performance of the model.

Given multiple views and a rough 3D model featuring noise-laden poses characterised by a set of Gaussian distributions $G$, we investigated how to apply gradient descent to minimise the residuals between GT pixels and the real view. Specifically, we adopted a joint optimisation strategy, tuning the parameters of all Gaussian distributions as well as the parameters of the camera. Such tuning enables $G$ to achieve a significant reduction in the photometric error at the target view position $T$. The corresponding equation is expressed as follows:
\begin{equation}
  \label{eq:GT}
  {G}^{*}, {T}^{*}=\underset{{G}, {T}}{\arg \min } \sum_{v \in N} \sum_{i=1}^{H W}\left\|\tilde{{C}}_{v}^{i}({G}, {T})-{C}_{v}^{i}({G}, {T})\right\|
  \end{equation}
where the height and width of a pixel are defined as $W$ and $H$ respectively.

In contrast to traditional methods, the camera poses for test views are accurately known and typically estimated within a unified coordinate system in conjunction with the training views. However, our scenes involves test views with unknown or noisy poses. We maintain the Gaussian model, trained on the training views, in a frozen state while directing our efforts towards optimising the camera pose for the test view. The primary objective of this optimisation process is to minimise the photometric discrepancy between the synthetic image and the actual test view, thereby enhancing the precision of the alignment assessment.

\subsubsection{Loss Function}
In order to improve the accuracy of image quality evaluation, we use SSIM\cite{SSIM} and its $\mathcal{L}_{ssim}$ as the optimization term, and add volumetric regularization $\mathcal{L}_{vol}$ to jointly optimize the learnable parameters and MLP. This process is designed to allow the model to more accurately capture the $\mathcal{L}_1$ loss between pixel colors.
The total loss function is given by the following formula: 
\begin{equation}
  \label{eq:L}
\mathcal{L}=\mathcal{L}_1+\lambda_{ssim}\mathcal{L}_{ssim}+\lambda_{vol}\mathcal{L}_{vol},
\end{equation}
where $\lambda_{ssim}$ is SSIM loss weight and $\lambda_{vol}$ is volumetric loss weight. This formula represents the structure of the entire loss function, which combines SSIM and $\mathcal{L}_{ssim}$ terms to measure image quality, while limiting the complexity of the model through $\mathcal{L}_{vol}$ terms.
In particular, volumetric regularized $\mathcal{L}_{vol}$ is calculated as:
\begin{equation}
  \label{eq:Lvol}
\mathcal{L}_{vol}=\sum_{i=1}^{N_{ng}}\mathrm{Prod}(s_{i})
\end{equation}
${N_{ng}}$ refers to the number of neural Gaussians in the scene, which is an important parameter that we use to represent image features in our model. $\mathrm{Prod}(\cdot )$ is the product of the vector values, which in our case is the product of the scale ${s_i}$ of each neural Gaussian. In this way, the volumetric regularization term encourages the neural Gaussians amounts in the model to remain small, and the overlap between them is reduced. Such the regularization strategy helps prevent overlapping and improves the accuracy of the assessment.

Bundle-adjusting not only significantly improves the accuracy of the pose by iteratively adjusting the camera pose parameters, but also ensures consistency in the reconstruction process and avoids pose distortion caused by camera movement.

\section{EXPERIMENTS}
\label{sec:Experiments}
\subsection{Experimental Setup}
\paragraph{Dataset.} 
In order to fully evaluate the effectiveness of our method, we conducted exhaustive experiments on a total of three datasets. These three datasets contain four different scenes to ensure the applicability and robustness of our method in different environments. Specifically, we selected two scenes (train and family) from the Tanks \& Temples dataset\cite{Tanks_and_temples} and one scene (hydrant) from the CO3D dataset \cite{CO3D}, which are widely used within the field of novel view synthesis and highly recognised in the industry for their high quality and diversity. Where the input images contain 301 images from the train scene, 150 images from the family scene and 68 images from the hydrant scene.

\begin{figure} [ht]
   \begin{center}
   \begin{tabular}{c} %% tabular useful for creating an array of images 
   \includegraphics[width=\textwidth,keepaspectratio]{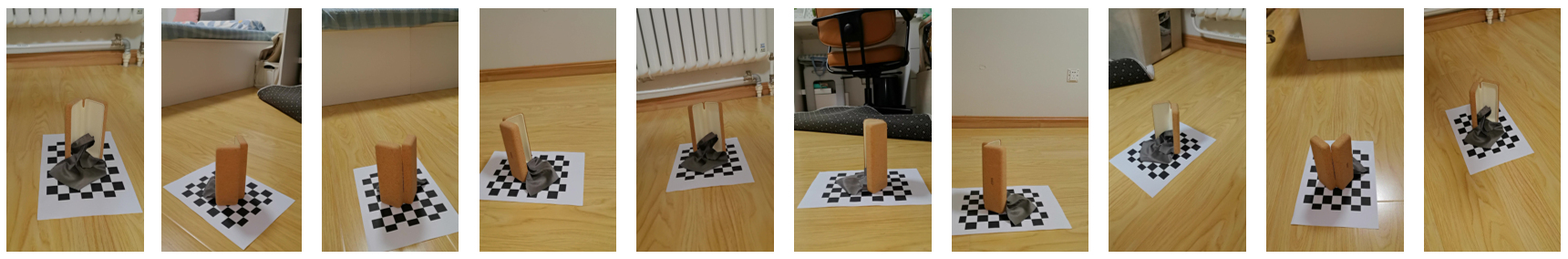}
   \end{tabular}
   \end{center}
   \caption[example] 
   { \label{fig:glasses} 
A number of representative input images of the glasses case have been selected for presentation. With these images, it can be observed that the input images not only contain a complete view of the glasses case, but also cover three different perspectives of its background: top, middle and bottom, thus providing a full range of information.}
   \end{figure}

Furthermore, we created a special dataset as shown in Fig.~\ref{fig:glasses}. The subject of this dataset is an glasses case and a lens wipe with folds placed on the floor of the room, and the background contains complexly rendered equipment such as a drainpipe. In order to capture detailed perspective information about these items, we used a handheld mobile phone to rotate and shoot video from three different angles: top, middle and bottom. This method allowed us to comprehensively capture the 3D structure and surface details of the entire scene. After data collection was completed, we randomly selected 102 frames from the video as input images.

For each scene, we created two folders in: a folder named images for storing the input images, and a blank folder named sparse for storing the subsequently generated sparse point cloud data. We performed preliminary feature extraction and matching operations on the input images stored in the images folder using SfM. In order to better manage these data, we created a new database file in .db format and set the images folder as the data source of this database. In the operation interface, we selected the options of “feature extract” and “feature matching” to start the process of feature extraction and matching. After completing these steps, we select the “reconstruction” option to start the reconstruction of the sparse point cloud. After a series of complex calculations and processing, the final sparse point cloud file will be saved in the previously created sparse folder for subsequent analysis and evaluation.

\paragraph{Metrics.} 
We chose three metrics to evaluate the experimental quality: PSNR, SSIM \cite{SSIM}, and LPIPS \cite{LPIPS}. PSNR, an engineering metric, quantifies the ratio of the maximum possible power of a signal to the power of the destructive noise that degrades its representation accuracy, thereby indicating pixel-level perception error. This metric treats all errors as noise without distinguishing between structural and non-structural distortions. SSIM, gauges both the distortion level within an image and the similarity between two images. In contrast to PSNR’s measurement of absolute error, SSIM operates as a perceptual model, aligning more closely with human visual perception. LPIPS is a metric for assessing image similarity that leverages a deep learning model to evaluate the perceptual differences between two images. LPIPS posits that even minute differences at the pixel level can lead to perceivable disparities to a human observer. Consequently, LPIPS employs a pre-trained deep neural network to extract image features and computes the distance between these features to determine the perceived similarity of the images.

\begin{figure} [ht]
   \begin{center}
   \begin{tabular}{c} %% tabular useful for creating an array of images 
   \includegraphics[width=\textwidth,keepaspectratio]{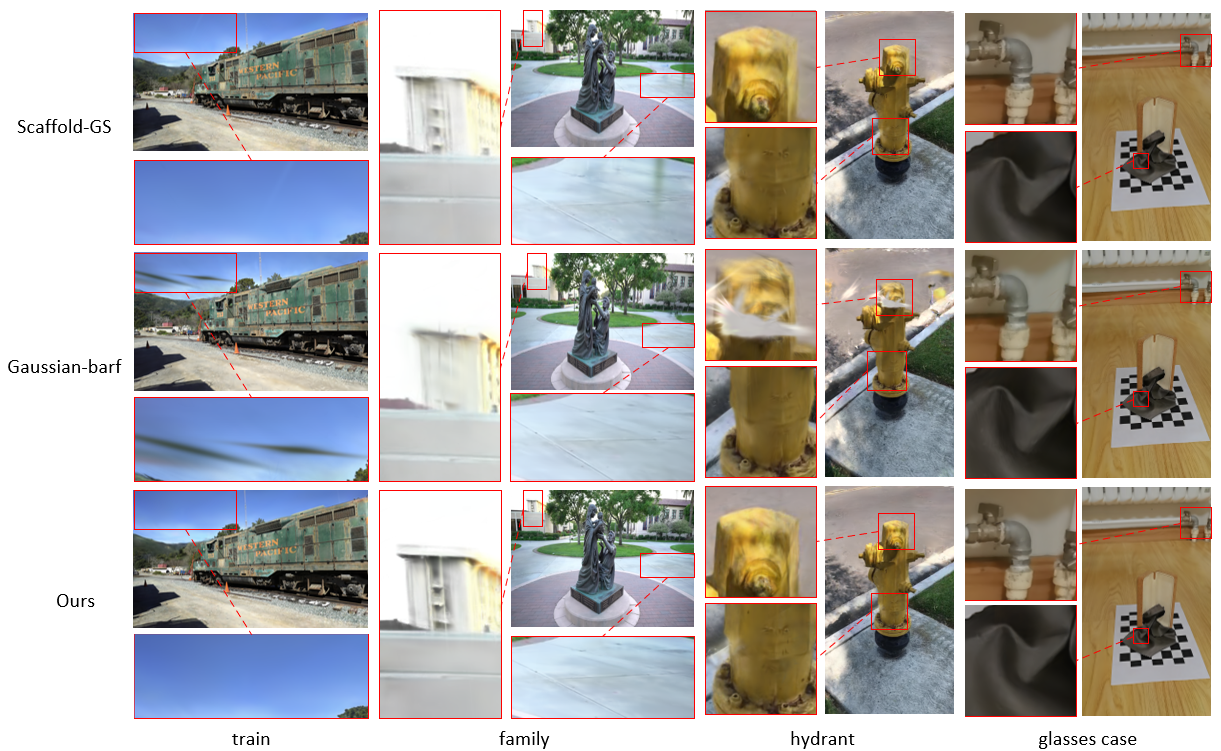}
   \end{tabular}
   \end{center}
   \caption[example] 
%>>>> use \label inside caption to get Fig. number with \ref{}
   { \label{fig:comparison} 
We selected two scenes from the Tanks \& Temples dataset (train and family), one scene from the CO3D dataset (hydrant), and one scene from the self-made dataset (Glasses case) to conduct experiments and validate the validity and generalization of our method.}
   \end{figure}

\paragraph{Baseline and Implementation.} 
After 30k iterations of training, we recorded the results of rendering the Tanks \& Temples, CO3D, and Glasses case datasets with our method, as shown in Fig.~\ref{fig:comparison}.
In our method, to enhance the precision of anchor localization, we deliberately configured each anchor to learn a set number of offsets, specifically $k=10$. The configuration facilitates the model’s ability to capture spatial information more effectively. Our method employs a streamlined two-layer MLP, with both layers equipped with the ReLU activation function to efficiently encode nonlinear characteristics. Furthermore, to maintain the model’s computational capacity, we standardized the number of hidden units across both layers to 32. Regarding the experimental setup of the loss function, we carefully fine-tuned the crucial loss weights: the SSIM loss weight, $\lambda_{ssim}$, was assigned a value of 0.2, while the volumetric loss weight, $\lambda_{vol}$, was set to 0.001.

We extract the translation vector and the quaternion vector from the camera poses generated during training and initialise these two vectors as well as the original poses. The iterative process will continue for 200 times in each loop of the test to ensure that the camera poses are optimised to the desired accuracy. Subsequently, we concatenate the initialised translation vector and the quaternion vector into a single vector that serves as the optimised camera pose so that we are able to obtain more accurate rendering results.

\begin{figure} [ht]
   \begin{center}
   \begin{tabular}{c} %% tabular useful for creating an array of images 
   \includegraphics[width=\textwidth,keepaspectratio]{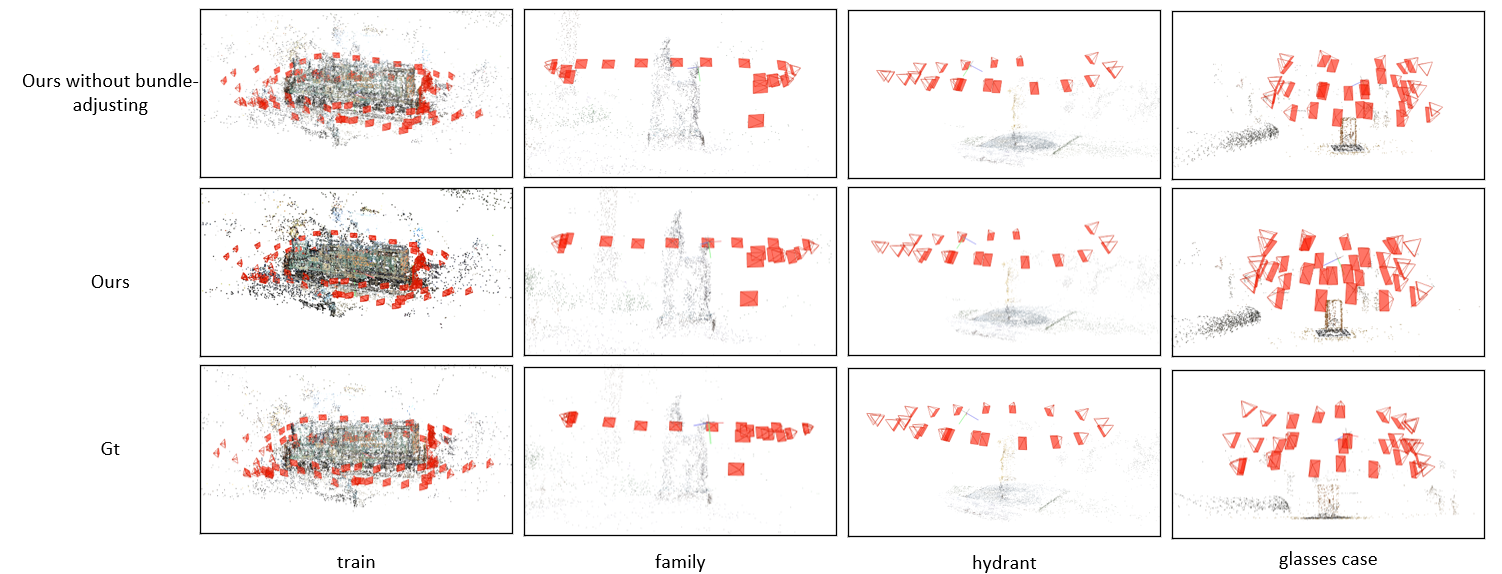}
   \end{tabular}
   \end{center}
   \caption[example] 
%>>>> use \label inside caption to get Fig. number with \ref{}
   { \label{fig:ablation} 
We show the results of our method to synthesise camera poses in four scenes. The first row is the camera pose by our method without bundle-adjusting; the second row is the camera pose by our method; and the third row is the camera pose of the ground truth(gt) images.}
   \end{figure} 
 
\subsection{Comparison with Baselines}
Table \ref{table:1} provides a detailed account of the experimental outcomes on three distinct datasets, wherein the performance of Scaffold-GS\cite{scaffold-gs}, Gaussian-barf\cite{FlowMap}, and our method is comparatively analyzed. The results indicate that our method consistently attains superior PSNR and SSIM metrics across the Tanks \& Temples, CO3D, and Glasses case datasets, signifying its notable edge in these benchmark evaluations.
As shown in Fig.~\ref{fig:comparison}, the rendered views produced by our method are notably richer in detail and exhibit greater sharpness, thereby reinforcing the outstanding capabilities of our method in the realm of image rendering. These findings comprehensively validate the efficacy of our method in augmenting both image quality and visual impact.

\begin{table}[H]
\scriptsize
\renewcommand{\arraystretch}{1.4}
\caption{\textbf{Quantitative metrics.} The subsequent section presents the outcomes of the numerical experiments conducted on the test view. The first row of the dataset indicate that no optimisation was performed for akin bit positions. The second row introduces bundle-adjusting optimisation, which is built upon the foundation of 3DGS\cite{3DGS}. The third row clearly demonstrates the superior performance of our proposed method.}  
\label{table:1} 
%\fontsize{12pt}{14pt}\selectfont{
\resizebox{\linewidth}{!}{
\begin{tabular}{c ccc ccc ccc }
\hline
  \multirow{2}{*} {\centering Method}         & \multicolumn{3}{c}{Tanks \& Temples}                                         & \multicolumn{3}{c}{CO3D}                                          & \multicolumn{3}{c}{Glasses case}       \\ 
& PSNR & SSIM & LPIPS & PSNR & SSIM & LPIPS & PSNR & SSIM & LPIPS   \\ \hline
Scaffold-GS\cite{scaffold-gs} & 24.92 & 0.852 & 0.163 & 25.51 & 0.855 & 0.203 & 31.55 & 0.930 & 0.188 \\ 
Gaussian-barf\cite{FlowMap} & 24.73 & 0.833 & 0.110 & 23.29 & 0.810 & 0.136 & 29.27 & 0.916 & 0.091 \\ 

Ours & 25.17 & 0.855 & 0.164 & 26.43 & 0.856 & 0.204 & 31.90 & 0.934 & 0.186\\ 
\hline

\end{tabular}
}
\end{table}

\subsection{Ablation Studies}
To ascertain the efficacy of incorporating point cloud features and pose optimization at anchor points, we carried out a suite of ablation studies. These experiments involved a comparative analysis of the baseline against our enhanced version, which includes two additional modules, on the Tanks\&Temples dataset. The detailed quantitative outcomes are presented in Table \ref{table:2}. The results show that, in contrast to the baseline, the integration of image features results in elevated PSNR and SSIM , coupled with a reduction in the LPIPS value, signifying the beneficial impact of image features on pose correction. Our proposed method is represented in the third row of the table, where, building on the second row, we have introduced a joint optimization module. This addition has led to an improvement in the PSNR and SSIM to 22.67 and 0.831, respectively, thereby substantiating the effectiveness of our method. The visualisation results of the image sparse reconstruction are shown in Fig.~\ref{fig:ablation}.

\begin{table}[!ht]
	\centering
	\caption{\textbf{Quantitative metric.} The numerical results obtained for the test images are shown below. The first row shows the results of the baseline, the second row presents the results after enhancing the image features based on the baseline, and the third row shows the experimental results of our proposed method. The experimental results show that our method achieves optimal results in all metrics.}
	\begin{tabular}{ccccccc}
    \toprule
    \multicolumn{1}{c}{\multirow{2}{*}{No.}} & \multirow{2}{*}{Baseline} & \multirow{2}{*}{Image Feature} & \multirow{2}{*}{Bundle-Adjusting} & \multicolumn{3}{c}{Tanks \& Temples (train)} \\
    \cmidrule{5-7}
    & & & & \multicolumn{1}{p{3.3em}}{PSNR} & \multicolumn{1}{p{3.3em}}{SSIM} & \multicolumn{1}{p{3.9em}}{LPIPS} \\
    \midrule
    1 & \checkmark & & & 22.18 & 0.813 & 0.196 \\
    2 & \checkmark & \checkmark & & 22.48 & 0.829 & 0.192 \\
    3 & \checkmark & \checkmark & \checkmark & 22.67 & 0.831 & 0.192 \\
    \bottomrule
\end{tabular}
	\label{table:2}%
\end{table}%

\section{CONCLUSION}
\label{sec:Conclusion}
In this work, we present a innovative mixed-beam adjusted 3D Gaussian model. The model’s primary advantage is its capability to produce high-fidelity renderings that are consistent across different viewpoints, along with precise optimization of poses. This is achieved through an integrated method that merges two distinct methods for extracting 3D representations: one based on image analysis and the other on neural network techniques. Our model can concurrently generate images that maintain viewpoint consistency and the corresponding camera poses within forward-facing scenes. To assess the efficacy of our proposed model, we performed a comprehensive series of experiments on a combination of real-world and synthetic datasets. The outcomes of these experiments confirm that our model delivers superior performance in rendering quality and pose optimization, underscoring its significant potential for practical applications.

\section*{ACKNOWLEDGMENTS}
This research was supported by the National Natural Science Foundation of China (No.62001033, No.U20A20163, No.62201066), and Beijing Municipal Education Commission Research Program (No.KZ202111232049, No.KM20\\
2111232014).


\begin{thebibliography}{10}

\bibitem{NeRF}
Mildenhall, B. et al., ``Nerf: Representing scenes as neural radiance fields for view
  synthesis,'' {\em Communications of the ACM}~{\bf 65}(1),  99--106 (2021).

\bibitem{NeRF--}
Wang, Z. et al., ``Nerf--: Neural
  radiance fields without known camera parameters,'' { https://arxiv.org/abs/2102.07064}  (2021).

\bibitem{3DGS}
Kerbl, B. et al., ``3d gaussian
  splatting for real-time radiance field rendering.,'' {\em ACM Trans.
  Graph.}~{\bf 42}(4),  139--1 (2023).

\bibitem{scaffold-gs}
Lu, T. et al., 
  ``Scaffold-gs: Structured 3d gaussians for view-adaptive rendering,'' in
  {\em IEEE Conf. Computer Vision and Pattern
  Recognit.(CVPR)}{\nolinebreak\hspace{0.1em}}, IEEE, pp.20654--20664 (2024).

\bibitem{SfM}
Schönberger, J.~L. and Frahm, J.-M., ``Structure-from-motion revisited,'' in
  {\em IEEE Conf. Computer Vision and Pattern
  Recognit.(CVPR)}{\nolinebreak\hspace{0.1em}},   IEEE, pp.4104--4113 (2016).

\bibitem{BARF}
Lin, C.-H. et al., ``Barf: Bundle-adjusting
  neural radiance fields,'' in {\em IEEE Int. Conf. Comput. Vision (ICCV)} ,  pp.5721--5731 (2021).

\bibitem{SPARF}
Truong, P. et al., ``Sparf: Neural
  radiance fields from sparse and noisy poses,'' in {\em IEEE Conf. Computer Vision and Pattern
  Recognit.(CVPR)}{\nolinebreak\hspace{0.1em}},   IEEE, pp.4190--4200 (2023).

\bibitem{FlowMap}
Smith, C. et al., ``Flowmap:
  High-quality camera poses, intrinsics, and depth via gradient descent,'' {https://arxiv.org/abs/2404.15259} (2024).

\bibitem{LightFR}
Levoy, M. and Hanrahan, P., ``Light field rendering,'' {\em Seminal Graphics
  Papers: Pushing the Boundaries, Volume 2}  (2023).

\bibitem{free}
Riegler, G. and Koltun, V., ``Free view synthesis,'' in {\em Proc. Eur. Conf. Comput. Vision
(ECCV)}{\nolinebreak\hspace{0.1em}}, Springer, pp.623--640 (2020).

\bibitem{DeepBF}
Hedman, P. et al., ``Deep blending for free-viewpoint image-based rendering,'' {\em ACM
  Transactions on Graphics (TOG)}~{\bf 37},  1 -- 15 (2018).

\bibitem{ScalableII}
Hedman, P. et al., ``Scalable
  inside-out image-based rendering,'' {\em ACM Transactions on Graphics
  (TOG)}~{\bf 35},  1 -- 11 (2016).

\bibitem{FreeVS}
Riegler, G. and Koltun, V., ``Free view synthesis,'' in {\em Proc. Eur. Conf. Comput. Vision
(ECCV)}{\nolinebreak\hspace{0.1em}},   Springer, pp. 623-640 (2020).

\bibitem{pixelNeRF}
Yu, A. et al., ``pixelnerf: Neural radiance
  fields from one or few images,'' in {\em IEEE Conf. Comput. Vision
and Pattern Recognit. (CVPR)}{\nolinebreak\hspace{0.1em}}, IEEE, pp.4578--4587 (2021).

\bibitem{LIRF}
Huang, X. et al., ``Local
  implicit ray function for generalizable radiance field representation,'' in
  {\em IEEE Conf. Comput. Vision
and Pattern Recognit. (CVPR)}{\nolinebreak\hspace{0.1em}}, IEEE, pp.97--107 (2023).

\bibitem{SC-NeRF}
Jeong, Y. et al.,
  ``Self-calibrating neural radiance fields,'' in {\em IEEE Int. Conf. Comput. Vision
(ICCV),}{\nolinebreak\hspace{0.1em}}, pp.5846--5854 (2021).

\bibitem{GARF}
Shi, Y. et al., ``Garf: Geometry-aware
  generalized neural radiance field,'' { https://arxiv.org/abs/2212.02280}
  (2022).

\bibitem{NoPe-NeRF}
Bian, W. et al., ``Nope-nerf:
  Optimising neural radiance field with no pose prior,'' in {\em IEEE Conf. Comput. Vision
and Pattern Recognit. (CVPR)}{\nolinebreak\hspace{0.1em}}, IEEE, pp.4160--4169 (2023).

\bibitem{PoRF}
Bian, J.-W. et al.,``Porf: Pose residual
  field for accurate neural surface reconstruction,'' {https://export.arxiv.org/abs/2310.07449}  (2023).

\bibitem{SSIM}
Wang, Z. et al.,``Image quality
  assessment: from error visibility to structural similarity,'' {\em IEEE
  Transactions on Image Processing}~{\bf 13}(4),  600--612 (2004).

\bibitem{Tanks_and_temples}
Knapitsch, A., et al., ``Tanks and temples:
  Benchmarking large-scale scene reconstruction,'' {\em ACM Transactions on
  Graphics (ToG)}~{\bf 36}(4),  1--13 (2017).

\bibitem{CO3D}
Reizenstein, J. et al., ``Common objects in 3d: Large-scale learning and evaluation of
  real-life 3d category reconstruction,'' in {\em IEEE Int. Conf. Comput. Vision
(ICCV)}{\nolinebreak\hspace{0.1em}},
   pp.10901--10911 (2021).

\bibitem{LPIPS}
Zhang, R., et al., ``The
  unreasonable effectiveness of deep features as a perceptual metric,'' in
  {\em IEEE Conf. Comput. Vision and Pattern Recognit. (CVPR)}{\nolinebreak\hspace{0.1em}}, IEEE, pp.586--595 (2018).

\end{thebibliography}
\end{document}